\documentclass[runningheads]{llncs}

\usepackage{eccv}

\usepackage{eccvabbrv}

\usepackage{graphicx}
\usepackage{booktabs}
\usepackage{subcaption} % for subtables
\usepackage{algorithm}
\usepackage{algpseudocode}
\usepackage{multirow}
\usepackage{pifont}
\usepackage{wrapfig,lipsum,booktabs}
\usepackage[accsupp]{axessibility}  % Improves PDF readability for those with disabilities.

\usepackage[pagebackref,breaklinks,colorlinks]{hyperref}
\usepackage{hyperref}

\usepackage{orcidlink}

\usepackage{xcolor}

\begin{document}

\title{TaskCLIP: Extend Large Vision-Language Model for Task Oriented Object Detection} 

\titlerunning{TaskCLIP: VLM for Task Oriented Object Detection}

\author{Hanning Chen\inst{1}\orcidlink{0000-0003-1956-5135}\thanks{This work was done during an Cisco internship.} \and
 Wenjun Huang\inst{1} \and Yang Ni\inst{1} \and Sanggeon Yun\inst{1} \and Yezi Liu\inst{1} Fei Wen\inst{2} Alvaro Velasquez\inst{3} Hugo Latapie\inst{4} Mohsen Imani\inst{1}\orcidlink{0000-0002-5761-0622}}

\authorrunning{Hanning Chen et al}

\institute{University of California Irvine, Irvine, CA 92697, USA \and
Texas A\&M University, College Station, TX 77843, USA \and
University of Colorado Boulder, Boulder, CO 80303, USA \and
Cisco, San Jose, CA 95134, USA  \\
\email{\{hanningc, m.imani\}@uci.edu}}

\maketitle
\begin{abstract}
Task-oriented object detection aims to find suitable objects for performing specific tasks. As a challenging task, it requires simultaneous visual data processing and reasoning under ambiguous semantics. Recent solutions are mainly all-in-one models. 
% To mitigate the ambiguity, the task is parsed into common visual attributes of suitable objects, by prompting the large language models. 
However, the object detection backbones are pre-trained without text supervision. Thus, to incorporate task requirements, their intricate models undergo extensive learning on a highly imbalanced and scarce dataset, resulting in capped performance, laborious training, and poor generalizability.
In contrast, we propose TaskCLIP, a more natural two-stage design composed of general object detection and task-reasoning object selection. Particularly for the latter, we resort to the recently successful large Vision-Language Models (VLMs) as our backbone, which provides rich semantic knowledge and a uniform embedding space for images and texts.
Nevertheless, the naive application of VLMs leads to suboptimal quality, due to the misalignment between embeddings of object images and their visual attributes, which are mainly adjective phrases. 
To this end, we design a transformer-based aligner after the pre-trained VLMs to recalibrate both embeddings. Finally, we employ a trainable score function to post-process the VLM matching results for object selection. 
Experimental results demonstrate that our TaskCLIP outperforms the DETR-based model TOIST in both accuracy ($+6.2\%$) and efficiency.
\noindent \keywords{Vision-Language Models \and Task-oriented Object Detection}
\end{abstract}

\section{Introduction} \label{sec:intro}
\begin{figure}[!t]%
\centering
\includegraphics[width=0.95\textwidth]{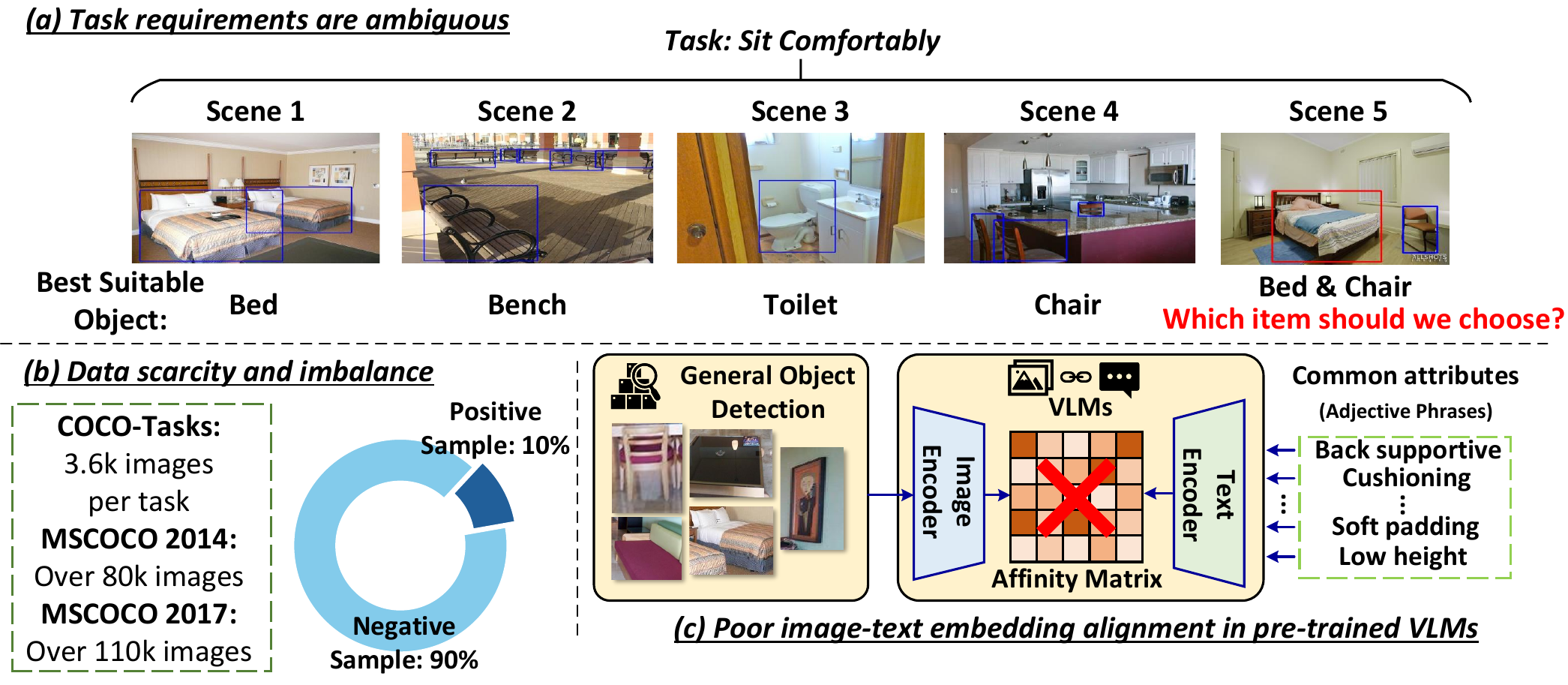}
\caption{(a) Ambiguity of task-oriented object detection. (b) Data scarcity and imbalance: suitable objects only take up a tiny portion of the total training samples. (c) Embedding misalignment when directly applying VLMs for object selection.} \label{fig: motivation}
\vspace{-2em}
\end{figure}

Object detection algorithms have seen tremendous progress on datasets like COCO~\cite{lin2014microsoft} and Pascal VOC~\cite{everingham2010pascal}, where they identify object instances of predefined categories in a scene~\cite{girshick2014rich, huang2024plug, yun2024hypersense}. 
However, in real-world applications, artificial intelligence is expected to handle more specific ``\textit{task-oriented object detection}''. To this end, researchers propose a new benchmark named COCO-Tasks~\cite{sawatzky2019object} that is much more challenging than traditional object detection tasks. Fig.~\ref{fig: motivation}(a) highlights that a simple, yet practical task never confines itself to a closed and predefined set of objects (or text prompts consisting of nouns). The inputs are in the form of affordance (e.g., ``\textit{Sit Comfortably}''), conditioning the selection of objects on their capability to fulfill a task. 
In particular, this presents a major challenge for prior object detection algorithms. The affordability as a prompt is intrinsically ambiguous, for example, the best suitable objects range from ``\textit{Bed}'' to the seemingly unrelated ``\textit{Toilet}'' for the same task “\textit{Sit Comfortably}”, depending on different scenes. 

Current solutions follow either a two-stage or a single-stage design to tackle this challenge. The former starts with regular object detection, followed by task-driven object selection, whereas the latter aims to achieve both sub-tasks with one single model. The two-stage design proposed in~\cite{sawatzky2019object, huang2024ecosense} applies a ranking of objects preference on top of objects extracted by traditional detectors. More recent one-stage designs~\cite{li2022toist} mainly resort to advanced object detection models such as the transformer-based DETR~\cite{carion2020end} as the backbone. 
Work~\cite{tang2023cotdet} further leverages Large Language Models (LLMs) to generate intermediate visual affordance features, which close the semantic gap between affordances that are mainly verb phrases and object types that are nouns. 

However, as intriguing as the all-in-one model sounds, the costs and difficulties behind it cannot be ignored. As pointed out by the previous solution~\cite{carion2020end}, transformers are considered data hungry, while obtaining data sets such as COCO-Tasks remains extremely difficult.
As shown in Fig.~\ref{fig: motivation}(b), the dataset also suffers from severe class imbalance, since only a handful of objects are relevant to the tasks concerned and the rest are labeled as negative samples.
In addition, all prior works require an explicit input of task-specific information to the object detection or selection model~\cite{sawatzky2019object, li2022toist}, fundamentally limiting their robustness against dataset domain shift or even slight changes in the affordance prompts. For instance, altering the task name to synonymous terms will cause a notable decline in the model's performance, even if the underlying characteristics of the task remain unchanged.
\textbf{This brings us to the question of whether it is optimal to train these specialized one-stage models for specific tasks from scratch, considering the limited data available in this domain. It also raises another important issue: how can we design more versatile computer vision models capable of addressing a variety of tasks?}

Our pursuit of solutions takes cues from the reasoning process of human beings, which is naturally multi-stage. Initially, they identify potential instances of objects and extract their visual features. During the process, humans parse the task, discerning the visual or functional attributes required of the candidate objects that are essential for task completion. Subsequently, they evaluated instances against these tasks-relevant attributes, selecting instances that align the most closely with the task requirements. Crucially, humans do not simply seek instances directly matching the task; instead, they engage in a step-by-step reasoning process that takes advantage of the common features of suitable objects as clues. This is similar in spirit to existing two-stage solutions. However, their performance is suboptimal mainly due to the bottleneck in the second stage, i.e., the association of object visual features and the semantics in task affordance. 

In this paper, we present TaskCLIP, a two-stage task-oriented object detection algorithm that tackles challenges in prior algorithms all at once, motivated by the recent success of large-scale Vision-Language Models (VLMs) on zero-shot transfer for language-supervised vision tasks. In the first stage, we perform general object detection~\cite{girshick2015fast, redmon2016you} for each scene. Meanwhile, we parse the task affordance, via LLMs\cite{brown2020language,ouyang2022training}, into several adjectives that describe common attributes of suitable objects. In our second stage, VLMs like CLIP~\cite{radford2021learning} and Flamingo~\cite{alayrac2022flamingo} serve as pre-trained backbones, which match image and text embeddings. The resulting affinity matrices then guide the selection of suitable objects. They enable the alignment of visual and text features by constructing a unified and high-quality embedding space, thereby resolving the bottleneck in current two-stage designs. More importantly, we avoid the complex customization, costly training, and poor generalizability of all-in-one models, where all parameters are learned from scratch and tuned specifically for a few tasks in COCO-Tasks with limited training samples. \textbf{Given that TaskCLIP addresses the task through the selection of items meeting visual characteristics, our model can be utilized for other tasks not included in the training group, provided they share the same fundamental attributes.}   
% An example of the object selection stage in TaskCLIP is illustrated in Fig.~\ref{fig: motivation}(c). Given input images, our model calculates an affinity matrix between objects and the text descriptions, aiming to maximize the similarity between the embeddings of real image-text pairs while minimizing the similarity of incorrect pairings.

However, incorporating VLMs like CLIP into our two-stage design is not trivial. Standard multimodal VLMs, while proficient in alignment, fall short of the reasoning capabilities required to select the most suitable objects. On one hand, these models primarily focus on aligning nouns with visual embeddings, while abstract concepts, such as \textbf{phrases with adjectives}, are poorly supported (as shown in Fig.~\ref{fig: motivation}(c)). On the other hand, large VLMs are trained on rich and diverse natural language annotations that contain not only nouns for objects. In response, we propose a re-alignment mechanism that blends seamlessly into the end-to-end model training. It reinstates the multimodal VLM's ability to encode adjectives within the shared embedding space and accurately associate them with the related visual embeddings.

In short, we argue that a two-stage model is indeed a more natural, generalizable, efficient, and effective design. Our contributions are summarized as follows:
\vspace{-1em}
\begin{itemize}
    \item Instead of training a multi-modal model from scratch, we propose the first VLM-based, two-stage, task-oriented object detection framework, leveraging the extensive semantic information from vision-language pretraining and its capability to support a calibrated joint vision-text embedding space. 
    \item We design a transformer-based aligner module to recalibrate the vision and text embeddings from VLMs, ensuring good matching between object visual features and adjective phrases from object common attributes. 
    \item To further mitigate the high false negative rate caused by imbalanced training samples, we specifically design a select-by-grouping mechanism. It ensures that all instances of the most suitable object are selected, by fully making use of the available knowledge from general object detectors. 
    \item Our framework provides natural task-driven reasoning and high-quality object detection, while also demonstrating higher training efficiency than previous works. In contrast to earlier DETR-based models, TaskCLIP attains similar accuracy on the COCO-Tasks dataset, yet exhibits significantly improved training efficiency and better generalizability. 
    %Compared to the DETR-based models, our TaskCLIP achieves up to 3.5\% improvement in mAP@0.5 on the COCO-Tasks dataset, requiring only a single RTX 4090 GPU to support both training and inference. 
\end{itemize}

\section{Related Work}
\begin{figure}[!t]%
\centering
\includegraphics[width=0.5\textwidth]{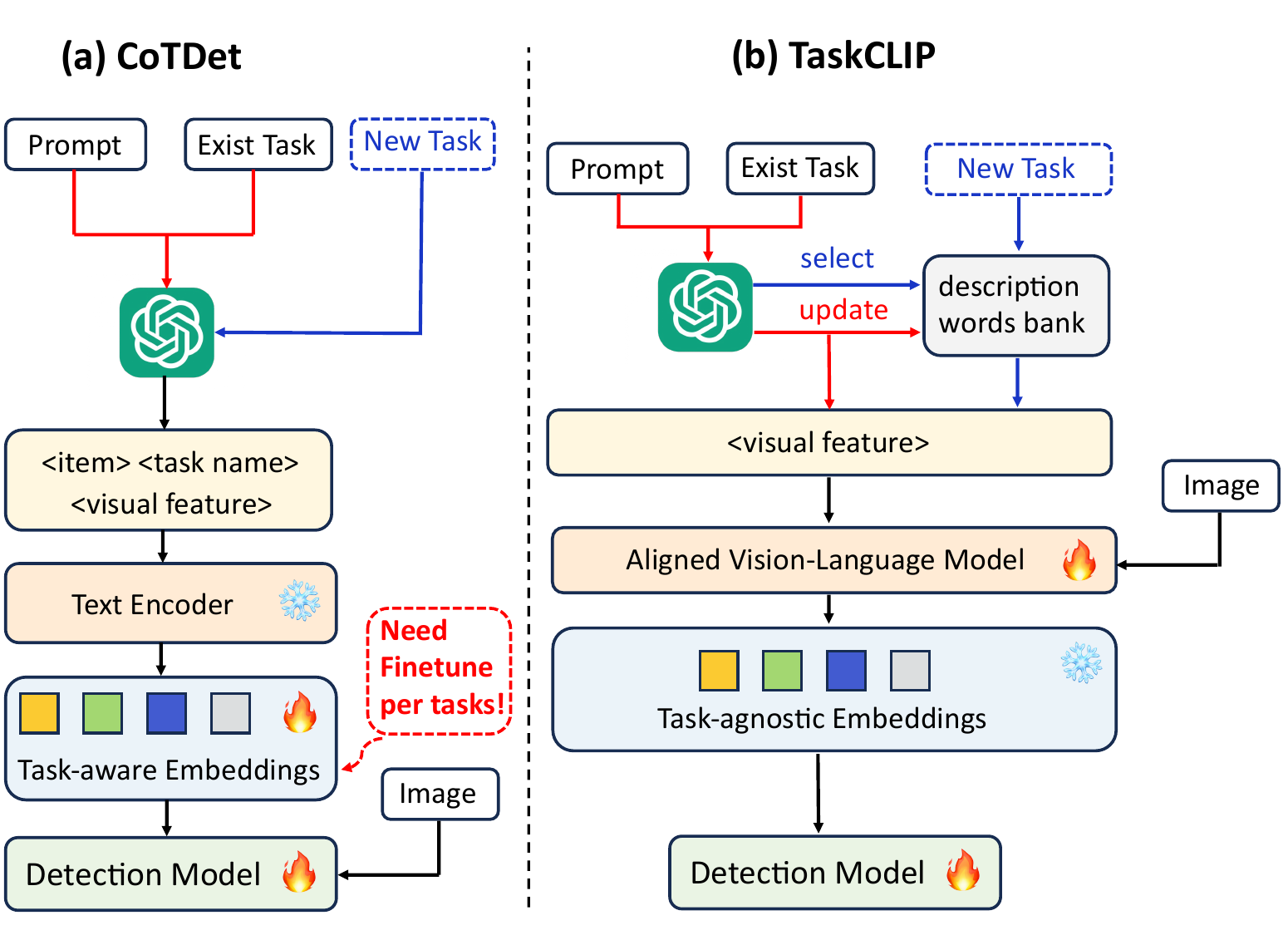}
\caption{(a) CoTDet~\cite{tang2023cotdet} workflow. (b). TaskCLIP (this work) workflow.} \label{fig: task_agnostic}
\vspace{-1em}
\end{figure}

\subsection{Task-oriented Object Detection}
This focuses on identifying the most suitable objects within an image to accomplish a specific task, such as serving wine. 
It requires the ability to reason and select objects based on a deep understanding of the task and the visual scene. 
% The reasoning usually starts with considering the common attributes of objects and their relationships within the scene. 
Thus, task-oriented object detection is much more complex than simple visual object detection~\cite{dvornik2018modeling,du2019visdrone,ren2015faster,liu2016ssd} and scene understanding~\cite{deng2021transvg,xiao2018unified,schon2021mgnet,cordts2016cityscapes, program, legal}.
The first task-driven object detection dataset, named COCO-Tasks, is proposed by a study~\cite{sawatzky2019object} based on MS COCO 2014~\cite{lin2014microsoft}. It also introduced a two-stage framework as a benchmark, which initially employs an existing object detection network to identify potentially suitable items, followed by gated graph neural networks (GGNN)~\cite{li2015gated} to model global inter-object relationship for object preference ranking.
Subsequent works, TOIST and CoTDet~\cite{li2022toist,tang2023cotdet}, are based on DETR~\cite{zhu2020deformable,kamath2021mdetr,carion2020end} and can be considered as single-stage frameworks. Compared to GGNN, the self-attention module in DETR-based models is more effective in terms of integrating text and visual information.  
However, training DETR-based models for task-oriented object detection faces multiple challenges due to the complex model structure and the large number of parameters. In contrast, our TaskCLIP resorts to the pertaining vision language and offers a more efficient alternative. A further problem is that while previous DETR-based models include text modalities, they do not take into account the model's ability to generalize. Consider CoTDet~\cite{tang2023cotdet}, as shown in Figure~\ref{fig: task_agnostic}.(a). For each task, after generating task-specific visual affordance and visual feature embeddings using the LLM (ChatGPT) and the text encoder (RoBERTa~\cite{liu2019roberta}), the model requires fine-tuning of these embedding vectors. A change in the task name results in a significant decrease in detection accuracy.

\subsection{Vision Language Model}
Extending pre-trained VLMs to support new applications has demonstrated their success in recent years. Among the most successful efforts, Flamingo~\cite{alayrac2022flamingo}, OpenAI CLIP~\cite{radford2021learning}, and OpenCLIP~\cite{cherti2023reproducible} have exhibited impressive performance in handling image-text matching due to their deep semantic knowledge and a comprehensive understanding of content that spans both modalities. These models have been applied successfully in downstream applications such as object detection~\cite{shi2022proposalclip}, action recognition~\cite{wang2021actionclip}, image captioning~\cite{mokady2021clipcap}, dense prediction~\cite{zhou2023zegclip}, disease prognosis~\cite{prognosis}, and semantic segmentation~\cite{liang2023open, medical_semantic, Li19}. However, existing CLIP-based algorithms primarily focus on matching image patches with nouns in text. Thus, additional calibration is necessary to facilitate the matching between image patches and visual attributes that are adjective phrases.
%\vspace{-1.5em}
\section{Method}
%\vspace{-0.5em}
\begin{figure*}[!t]%
\centering
\includegraphics[width=0.95\textwidth]{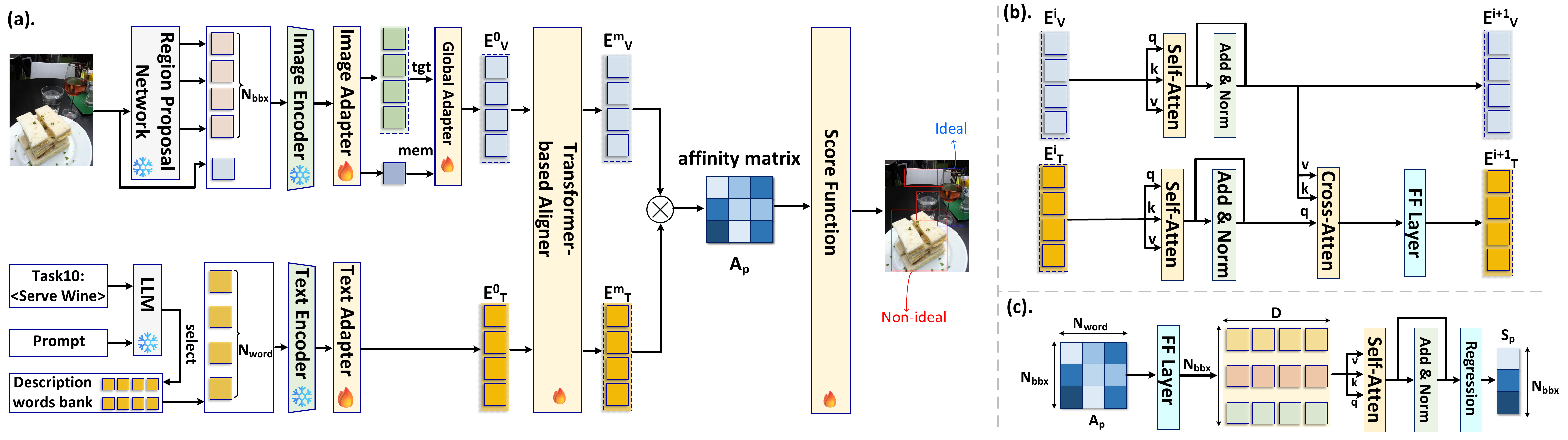}
\caption{TaskCLIP model architecture: (a) The overall framework of the design. (b) The architecture of the aligner module. (c) Detailed design of the score function}\label{fig: model_arch}
%\vspace{-2em}
\end{figure*}

\subsection{Task visual attributes preparation}

% \begin{figure*}
%       \centering
%       \includegraphics[width=0.9\linewidth]{Figure/GPT Prompt.pdf}
%       \caption{TaskCLIP LLM prompt.} \label{fig:gpt_flow}
% \end{figure*}

\begin{table*}[]
\caption{TaskCLIP LLM prompt.}
\centering
\label{tab:LLM_prompt}
\resizebox{0.7\textwidth}{!}{
\begin{tabular}{lll}
\toprule
\multicolumn{3}{c}{\textbf{(a) Description words bank preparation} } \\
\multicolumn{3}{l}{\begin{tabular}[c]{@{}l@{}}\textbf{User}: Please list items to solve the task of \textless{}Task Name\textgreater{}?\\ \textbf{LLM response}: {[}item1, item2, ...{]}\\ \textbf{User}: Summarize visual featuresof the {[}item1,item2,...{]} for \textless{}Task Name\textgreater{}.\\ \textbf{LLM response}: {[}feature1, feature2, ...{]}\end{tabular}} \\ \midrule
\multicolumn{3}{c}{\textbf{(b) Synonym task generation} } \\
\multicolumn{3}{l}{\begin{tabular}[c]{@{}l@{}}\textbf{User}: Please generate a synonym for \textless{}Task Name\textgreater\\ \textbf{LLM response}: \textless{}Task Name Synonym\textgreater{}\end{tabular}} \\ \midrule
\multicolumn{3}{c}{\textbf{(c) Visual features selection}} \\
\multicolumn{3}{l}{\begin{tabular}[c]{@{}l@{}}\textbf{User}: Please select 20 visual feature words from words bank for \textless{}Task Name\textgreater\\ \textbf{LLM response}: {[}feature1, feature2, ....{]}\end{tabular}} \\ \bottomrule
\end{tabular}}
\vspace{-1em}
\end{table*}

To close the semantic and reasoning gap between task affordance and suitable object class, we utilize an LLM to produce task-relevant visual common attributes, as is shown in Fig.~\ref{fig: task_agnostic}(b). In Table~\ref{tab:LLM_prompt}.(a), we delineate the prompts used to guide the LLM in generating the appropriate visual attributes for each task. To illustrate this process, we consider task 10 \textbf{(Serve wine)} as an example. For task 10, the procedure unfolds as follows:

\noindent (1). We initiate the process by querying the LLM: ``\textit{Please list items to solve the task of \textbf{Serve wine}?}'' 
The response generated by the LLM encompasses several real-life items, such as ``wine glass'', ``cup'', ``stemware'', and others.

\noindent (2). Building on this initial response, we proceed to the second prompt: ``\textit{Summarize the visual features of the [wine glass, cup, stemware, ...] for \textbf{serve wine}.}'' 
It is designed to guide the LLM in summarizing the visual characteristics of the mentioned items specifically in the context of the ``serve wine'' task.

\noindent (3). Ultimately, we obtain the common visual attributes of objects that can afford this task from LLM responses, such as ``narrow rim'' and ``transparency''. In total, we generate $\mathbf{N_{word}}=20$ common attributes.

\noindent (4). We save these generated common attributes in the description words bank. As depicted in Figure~\ref{fig: task_agnostic}, when new tasks arise, the LLM will pick the most suitable attribute from the description words bank. Since TaskCLIP is trained by aligning item images with visual feature description words, even if the task name changes, as long as the intrinsic attributes of the new task can be articulated using the words from the bank, our model remains effective. Hence, TaskCLIP exhibits superior task generalizability compared to prior works.  

\vspace{-1em}

\subsection{Text and visual embedding vector generation}
\vspace{-0.5em}
Meanwhile, we generate $\textbf{$N_{bbx}$}$ bounding boxes for all objects in the scene as is shown in Fig.~\ref{fig: model_arch}(a). During the training process, we directly use ground truth bounding boxes, whereas, for inference, we use a pre-trained object detection network (such as Faster R-CNN~\cite{girshick2015fast} or YOLOv8~\cite{jocher2020ultralytics}). Notice that in the basic setting of our framework during inference, the object detection network is only responsible for generating bounding boxes for all items in the image. 
% In other words, we use the front-end object detection network merely as a region proposal network. 
We will introduce a grouping mechanism in Section~\ref{sec:class_forward} to further leverage object classification information and guide our model's final decision. 
Based on the coordination of the bounding box, we extract $N_{bbox}$ image patches. After generating $N_{word}$ common visual attributes and $N_{bbox}$ image patches, we pass them into pre-trained VLMs (such as OpenAI CLIP~\cite{radford2021learning}) to generate text and image embeddings as is shown in Fig.~\ref{fig: model_arch}(a). Here we use $L_{word}$ and $L_{bbox}$ to represent the lists of visual feature words and image patches. The lengths of the lists are $N_{word}$ and $N_{bbox}$, respectively. The computation process is summarized as:
\begin{equation}
    \mathbf{E^0_V} = CLIP_{image}(L_{bbox})
\end{equation}
\begin{equation}
    \mathbf{E^0_T} = CLIP_{text}(L_{word})
\end{equation}
Suppose the embedding dimension is $\textbf{D}$, the shape of the generated vision embedding matrix and text embedding matrix is $\textbf{N}_{bbox}$$\times$$\textbf{D}$ and $\textbf{N}_{word}$$\times$$\textbf{D}$. To better introduce the new knowledge into the vision language models, we also add vision and text adapter~\cite{gao2024clip} after the CLIP model.
\begin{equation}
    \mathbf{E^0_V} = (1 - \alpha) \times \mathbf{E^0_V} + ReLU(\mathbf{E^{0T}_V}\mathbf{W_1^v})\mathbf{W_2^v}
\end{equation}
\begin{equation}
    \mathbf{E^0_T} = (1 - \beta) \times \mathbf{E^0_T} + ReLU(\mathbf{E^{0T}_T}\mathbf{W_1^t})\mathbf{W_2^t}
\end{equation}
Here $W_{1}^{v}$, $W_{2}^{v}$ and $W_{1}^{t}$, $W_{2}^{t}$ are the weights of bottleneck linear layers for vision and text adapter networks. $\alpha$ and $\beta$ are hyper-parameters, here we choose 0.3.

\subsection{Global attention}
To gain a clearer understanding of the role of each bounding box item within the overall scene, we initially encode the input image using the CLIP vision encoder. Subsequently, we use the embedding of the bounding box list ($E^0_V$) as the target and the image embedding as the memory. After the cross-attention module, we get the bounding box embedding with global information.

\subsection{Vision and text embedding space recalibration}
Although OpenAI CLIP and its derivative works~\cite{radford2021learning,zhong2022regionclip} achieve high accuracy in matching text with image patch, they focus on \textbf{noun} (such as ``wine glass'' and ``cup'') instead of \textbf{visual attributes} (such as ``transparency'' and ``narrow rim''). To match the image patches with task-specific visual attribute words, it is pivotal to re-calibrate the embedding space of vision and text. Therefore we design a transformer-based \textbf{aligner} module, next to the pre-trained VLM. In Fig.~\ref{fig: model_arch}(a), we pass the vision embedding matrix and text embedding matrix into a multi-layer aligner module to generate the new re-calibrated vision and text embedding matrix, $\textbf{E}^{\textbf{m}}_{\textbf{V}}$ and $\textbf{E}^{\textbf{m}}_{\textbf{T}}$ respectively. Here \textbf{m} represents the number of layers in aligner module. In Fig.~\ref{fig: model_arch}(b), we present the network architecture of each layer of the aligner module. There are multiple layers of aligner module where each layer has two self-attention modules, one cross-attention module, and one feed-forward network layer. For the $\textbf{i}^\textbf{th}$ aligner module layer, the input ($\textbf{E}^{\textbf{i}}_{\textbf{V}}$ and $\textbf{E}^{\textbf{i}}_{\textbf{T}}$) is the vision and text embedding matrix coming from last aligner module layer. At each layer, both vision and text embedding matrix will first go through a self-attention layer. The purpose of the self-attention layer is to learn the most optimal affordance that could be used to solve this task. For example, as is shown in Fig.~\ref{fig: motivation}(a) Scene 5, to solve the task of \textbf{sit comfortably} we prefer to choose chair over bed. After self-attention layer, we use a cross-attention layer to recalibrate the text embedding and vision embedding. The last part of the aligner network is a fully connected feed forward layer where the text embedding matrix will pass through. Like previous works that try to fine-tune pre-trained vision-language model~\cite{gao2023clip,goyal2023finetune}, in our aligner module, we focus on the text embedding vectors transformation. Therefore, we only pass text embedding matrix through the last feed forward layer.    

\subsection{Score function and training}
After m transformer aligner layers' processing, we get the recalibrated image vision embedding ($\textbf{E}^\textbf{m}_\textbf{V}$) and text embedding ($\textbf{E}^\textbf{m}_\textbf{T}$). In \cref{fig: model_arch}(a), we perform matrix to matrix multiplication between $\textbf{E}^\textbf{m}_\textbf{T}$ and $\textbf{E}^\textbf{m}_\textbf{T}$ to generate the predicted affinity matrix $\textbf{A}_{p}$. The computation could be summarized as:
\begin{equation} \label{eq: affinity}
    \mathbf{A}_{\mathbf{p}} = \mathbf{E}^\mathbf{m^T}_{\mathbf{V}} \times \mathbf{E}^\mathbf{m}_{\mathbf{T}}
\end{equation}
The shape of the affinity matrix $\textbf{A}_{\textbf{p}}$ is $N_{bbx} \times N_{word}$.
In traditional CLIP, to perform the zero-shot image classification, we will directly apply softmax function over affinity matrix to get the predict label. In task-oriented object detection, since we need to find the most-optimal items, further processing is necessary.
Regarding the meaning of $\textbf{A}_{\textbf{p}}$[i][j], it represents the score that bounding box i have visual attribute j. To make the final decision of choosing the most optimal item that could be treated as the ideal task affordance, we pass the affinity matrix into a score function. In Fig.~\ref{fig: model_arch}(c), we present the score function model architecture. It will first encode each bounding box's visual feature score vector into a higher dimensional vector. In Fig.~\ref{fig: model_arch}(c), we represented the encoded new matrix as $\textbf{H}_\textbf{p}$ whose dimension is $N_{bbox} \times D'$. This encoding process is:
\begin{equation}
    \mathbf{H}_\mathbf{p}[j] = MLP(\mathbf{A}_\mathbf{p}[j]) \quad where \quad j \in [0,N_{bbox})
\end{equation}
Then we pass $\textbf{H}_\textbf{p}$ into a self-attention layer. Here, we incorporate a self-attention layer because we aim to learn the relative importance of different items, enabling us to select the most suitable items for this task. Following the self-attention layers, we employ multiple fully connected layers for regression, ultimately producing the score vector, ($\Vec{S}_{p}$). The shape of $\Vec{S}_{p}$ is $1 \times N_{bbox}$ while each element of $\Vec{S}_{p}$ has value ranging from 0 to 1. For inference, we will set a threshold value ($\delta$) and we have:
\begin{equation}
    \mathbf{P}[i] = \left\{
    \begin{array}{ll}
        1 & \text{if } \Vec{S}_{p}[i] \geq \delta, \\
        0 & \text{else }.
    \end{array}
\right.
\end{equation}
Here $\textbf{P}[i]$ is the prediction of whether the item of bounding box i is an ideal affordance of this task.
During training, we will have the ground truth score vector $\Vec{S}_{g}$. For the $i^th$ element of $\Vec{S}_{g}$, if it can solve the task then its value is 1 otherwise its value is 0. The loss \textbf{L} during the back propagation is:
\begin{equation}
    \mathbf{L} = MSELoss(\Vec{S}_{g}, \Vec{S}_{p})
\end{equation}
Here $MSELoss$ is the mean squared error loss function~\cite{james1992estimation}. After getting the loss, we perform the back propagation and conduct end to end training of CLIP adapter, global attention layer, transformer aligner module, and score function as is shown in Fig.~\ref{fig: model_arch}(a). Unlike end-end to training style of DETR-based works, our framework will not touch object detection network and vision-language model. Reuse vision-language pretraining knowledge make our framework has higher training efficiency when compared with previous works, as is shown in section~\ref{sec:experiment}.

\begin{table*}[]
\centering
\caption{Original COCO-Tasks dataset task name and corresponding synonym task name from ChatGPT. The synonym task name is in the parentheses.}
\label{tab:dataset_and_experiment}
\resizebox{\textwidth}{!}{
\begin{tabular}{lllll}
\toprule
task1: step on (trend on)              &  & task2: sit comfortably (lounge easily)  &  & task3: place flowers (arrange blooms) \\
task4: get potatoes out of fire (retrieve tubers from heat) &  & task5: water plant (irrigate vegetation) &  & task6: get lemon out of tea (remove citrus from brew) \\
task7: dig hole (excavate pit)         &  & task 8: open bottle of beer (uncap ale) &  & task9: open parcel (unwrap package)   \\
task10: serve wine (serve a cocktail)  &  & task11: pour sugar (dispense sweetener) &  & task12: smear butter (spread cream)   \\
task13: extinguish fire (douse flames) &  & task 14: pound carpet (beat rug)        &  &                                       \\ \bottomrule
\end{tabular}}
% \vspace{-2em}
\end{table*}

\subsection{Select-by-grouping mechanism} \label{sec:class_forward}
When using existing object detection network, such as Faster R-CNN, we not just get the bounding box (bbox) information but also the bbox COCO class. To assist us find all suitable affordance, here we propose the select-by-grouping mechanism. Suppose in an image, we find a total of N bbox. After the score function, we find that the $i^{th}$ bbox's regression score is higher than this task's threshold. Considering the bbox quality, it is possible that we omit some items which are the same COCO class of the $i^th$ bbox. 
In most cases, the same class items show a high chance to be ideal task affordance. To compensate this false-negative drawback, our select-by-grouping mechanism will check all other bbox and see if the following scenario satisfied:
% \begin{equation} \label{eq:forward}
%     \mathbf{P}[j] = 1 \quad if \quad \mathbf{class}[j] = \mathbf{class}[i] \quad and \quad \mathbf{conf}[j] > \beta \quad and \quad \mathbf{conf}[i] > \beta
% \end{equation}
\begin{equation} \label{eq:forward}
\begin{split}
    \mathbf{P}[j] = 1 \quad &\text{if} \quad \mathbf{class}[j] = \mathbf{class}[i] \\
    &\text{and} \quad \mathbf{conf}[j] > \beta \\
    &\text{and} \quad \mathbf{conf}[i] > \beta
\end{split}
\end{equation}

Here $\textbf{P}$ is the final prediction to see if bbox i is an ideal affordance. The shape of $\textbf{P}$ is $1 \times N$. The \textbf{class} and \textbf{conf} are also two $1 \times N$ vectors which represent the bbox COCO class type and class confidence. $\beta$ is a hyperparameter that identifies whether the front-end detection network has sufficient confidence in the bounding box classification result. In our experiment, we set it to 0.8. The meaning of equation~\ref{eq:forward} is that suppose we found the $j^th$ bbox has the same COCO class with the $i^th$ bbox with high confidence, we also set the $j^{th}$ bbox item as an ideal affordance. As we shown in Fig.~\ref{fig: motivation}(b), since the samples of category 1 items are much less than category 0, our original model has high true positive and false negative rate. The select-by-grouping mechanism therefore will reduce the false negative prediction. 

\section{Experiments} \label{sec:experiment}
\subsection{Datasets and Metric}

% \begin{table*}[]
% \caption{COCO-Tasks dataset task name.}
% \label{tab:dataset_and_experiment}
% \resizebox{\textwidth}{!}{
% \begin{tabular}{lllllll}
% \toprule
% task1: step on &  & task2: sit confortably &  & task3: place flowers &  & task4: get potatoes out of fire \\
% task5: water plant &  & task6: get lemon out of tea &  & task7: dig hole &  & task 8: open bottle of beer \\
% task9: open parcel &  & task10: serve wine &  & task11: pour sugar &  & task12: smear butter \\
% task13: extinguish fire &  & task 14: pound carpet &  & \multicolumn{1}{c}{-} &  & \multicolumn{1}{c}{-} \\ \bottomrule
% \end{tabular}}
% \end{table*}

% \begin{table}[]
% \caption{{Experiment setup and model comparison.}}
% \label{tab:model_compare}
% \resizebox{0.45\textwidth}{!}{
% \begin{tabular}{c|cc}
% \toprule
% \textbf{Algorithm}                                                              & DETR-based methods  & TaskCLIP              \\ \midrule
% \textbf{Platform}                                                                        & $\times$8 A100 GPUs & $\times$1 RTX4090 GPU \\ \hline
% \textbf{Training epoch}                                                                  & 30 epochs           & 20 epochs             \\ \hline
% \textbf{\begin{tabular}[c]{@{}c@{}}Vision-Language \\ Pretraining\end{tabular}} &         \ding{55}            &              \ding{51}         \\ \hline
% \textbf{\begin{tabular}[c]{@{}c@{}}\# Trainable \\ Parameters\end{tabular}}     & Large               & Small                 \\ \bottomrule
% \end{tabular}}
% \end{table}

\noindent \textbf{Dataset} 
We conduct experiments on the COCO-Tasks dataset \cite{sawatzky2019object}, which is derived from MS COCO2014 \cite{lin2014microsoft}.  
The main difference between COCO-Tasks dataset and MS COCO2014 lies in the labeling of each item. 
In COCO-Tasks, the objective is to detect the most suitable items applicable to solve specific tasks.
The COCO-Tasks dataset comprises a total of 14 tasks, each consisting of 3600 training images and 900 test images. 
Within each image, the labels of bounding boxes can either be categorized as category 0 (not ideal affordance) or category 1 (ideal affordance). 
It's important to note that the same item (e.g., a bed) may be classified into opposite categories (ideal affordance vs. not ideal affordance) for solving the same task (e.g., sitting comfortably) across different images, as illustrated in Fig.~\ref{fig: motivation}(c).
\newline \noindent \textbf{Task-agnostic experiment}
To evaluate the model's ability to generalize, we not only used the 14 original tasks from the COCO-Tasks dataset but also performed a task-agnostic experiment. Initially, referencing \cref{tab:LLM_prompt}.(b), for each original task in the COCO-Tasks dataset, we requested ChatGPT to generate synonymous task words, as displayed in \cref{tab:dataset_and_experiment}. For instance, the synonym task for \textit{\textbf{Serve wine}?} is \textit{\textbf{Serve a cocktail}?}. These two tasks share common affordances (such as wine glasses), so an effective generalization model trained on the original COCO-Tasks dataset should be capable of addressing the new task.
\newline \noindent \textbf{Metric} We use the AP@0.5 metric for task-oriented detection. 
Then we compute the mAP@0.5 by averaging the AP@0.5 accuracy across all 14 tasks. 
To keep consistency with previous studies~\cite{sawatzky2019object,li2022toist}, we only report accuracy results for predictions related to suitable affordances (category 1).  
\vspace{-1.5em}
\begin{table*}[t]
\caption{{Experiment setup and model comparison.}}
\centering
\label{tab:model_compare}
\resizebox{0.95\textwidth}{!}{
\begin{tabular}{c|c|c|c|c}
\toprule
         \textbf{Algorithm} & \textbf{Platform}       & \textbf{Training epoch} & \textbf{\begin{tabular}[c]{@{}c@{}}Vision-Language \\ Pretraining\end{tabular}}  & \textbf{\begin{tabular}[c]{@{}c@{}}\# Trainable \\ Parameters\end{tabular}} \\ \midrule
\multirow{2}{*}{\begin{tabular}[c]{@{}c@{}}DETR-based\\ methods~\cite{tang2023cotdet,li2022toist}\end{tabular}} &
  \multirow{2}{*}{$\times$8 A100 GPUs} &
  \multirow{2}{*}{30 epochs} &
  \multirow{2}{*}{\ding{55}} &
  \multirow{2}{*}{Large} \\
         &                &                &                        &               \\
         \hline
TaskCLIP & $\times$1 RTX4090 GPU & 20 epochs      & \ding{51} & Small          \\ \bottomrule
\end{tabular}
}
\end{table*}
\subsection{Implementation Details}
\vspace{-0.5em}
For the object detection network, we experiment with both Faster R-CNN~\cite{ren2015faster} and YOLOv8~\cite{Jocher_Ultralytics_YOLO_2023}.
The implementation of Faster R-CNN utilizes the X101-FPN configuration in Detectron2~\cite{wu2019detectron2}. As for YOLOv8, we employ the YOLOv8x configuration. 
Regarding the large vision language model (VLM), we employ OpenCLIP~\cite{cherti2023reproducible,radford2021learning}. Specifically, the vision transformer encoder configuration is ViT/H and the text encoder is RoBERTa~\cite{liu2019roberta}. We implement transformer aligner module and score function using PyTorch~\cite{paszke2019pytorch}. 
The model was trained for 20 epochs with an initial learning rate of 1e-6, employing AdamW~\cite{loshchilov2017decoupled} as the optimizer.
We generated visual feature attributes for each task using OpenAI ChatGPT~\cite{lund2023chatting}.

In Table~\ref{tab:model_compare}, we present a high-level comparison of experimental settings between TaskCLIP and previous DETR-based models, including TOIST and CotDet~\cite{li2022toist,tang2023cotdet}.  
Compared to transformer-based approaches, our training model is significantly smaller as we efficiently leverage pre-trained large VLM. Unlike training everything from scratch, our main training emphasis lies in recalibrating the vision and text embedding space.  Without employing any additional training assistance methods, such as knowledge distillation~\cite{gou2021knowledge}, our entire framework can be trained and deployed on \textbf{a single NVIDIA RTX 4090 GPU}.

\begin{table*}[t]
\centering
\caption{Comparison with previous methods. Values in parentheses indicate results from task-agnostic experiments.}
\label{tab:comparison}
\resizebox{1\textwidth}{!}{%
\begin{tabular}{lcccccc}
\toprule
\textbf{Task (AP@0.5(\%))} & \textbf{GGNN~\cite{sawatzky2019object}} & \textbf{TOIST~\cite{li2022toist}} & \textbf{TOIST$\dag$~\cite{li2022toist}} & \textbf{CoTDet~\cite{tang2023cotdet}} & \textbf{TaskCLIP} & \textbf{TaskCLIP*} \\ \midrule
task1  & 36.6 (-) & 44.0 (-) & 45.8 (-) & 58.9 (56.2) & 48.4 (47.6) & 52.1 (50.9) \\
task2  & 29.8 (-) & 39.5 (-) & 40.0 (-) & 55.0 (53.1) & 44.6 (44.1) & 47.4 (46.6) \\
task3  & 40.5 (-) & 46.7 (-) & 49.4 (-) & 51.2 (48.2) & 52.4 (52.4) & 54.1 (54.1) \\
task4  & 37.6 (-) & 43.1 (-) & 49.6 (-) & 68.5 (57.3) & 60.6 (60.6) & 62.5 (62.5) \\
task5  & 41.0 (-) & 53.6 (-) & 53.4 (-) & 60.5 (54.9) & 54.4 (54.4) & 56.9 (56.9) \\
task6  & 17.2 (-) & 23.5 (-) & 26.9 (-) & 47.7 (35.0) & 31.5 (31.5) & 34.0 (34.0) \\
task7  & 43.6 (-) & 52.8 (-) & 58.3 (-) & 76.9 (50.5) & 67.7 (67.7) & 69.9 (69.9) \\
task8  & 17.9 (-) & 21.3 (-) & 22.6 (-) & 40.7 (3.6) & 18.6 (18.6) & 19.2 (19.2) \\
task9  & 21.0 (-) & 23.0 (-) & 32.5 (-) & 47.4 (20.1) & 39.4 (39.4) & 41.4 (41.4) \\
task10 & 40.6 (-) & 46.3 (-) & 50.0 (-) & 66.5 (58.5) & 55.7 (55.7) & 58.1 (58.1) \\
task11 & 22.3 (-) & 33.1 (-) & 35.5 (-) & 41.9 (30.6) & 38.1 (38.1) & 39.8 (39.8) \\
task12 & 28.4 (-) & 41.7 (-) & 43.7 (-) & 48.3 (47.3) & 47.6 (47.6) & 49.8 (49.8) \\
task13 & 39.1 (-) & 48.1 (-) & 52.8 (-) & 61.7 (9.4) & 46.3 (46.3) & 48.7 (48.7) \\
task14 & 40.7 (-) & 52.9 (-) & 56.2 (-) & 71.4 (58.4) & 67.4 (67.4) & 69.8 (69.8) \\
\midrule
mean (mAP@0.5(\%)) & 32.6 (-) & 41.3 (-) & 44.1 (-) & 56.9 (40.9) & 48.1 (\textbf{47.9}) & 50.3 (\textbf{50.1}) \\
\bottomrule
\multicolumn{7}{l}{\small *TOIST$\dag$ represents the model with noun-pronoun distillation.}\\
\multicolumn{7}{l}{\small *TaskCLIP* is our method with a select-by-grouping mechanism, where we also apply different optimized thresholds for each task.}
\end{tabular}}
\end{table*}

\subsection{Comparisons with Previous Works}

In Table~\ref{tab:comparison}, we present a comparison of TaskCLIP's performance with previous works on the COCO-Task dataset. Here we present results for both original COCO-Task dataset tasks' results and synonym tasks' result (in the parentheses).  
We utilize Faster R-CNN as the front-end object detection network to identify bounding boxes (bbox) for each item. 
The transformer aligner module between OpenAI CLIP and the score function comprises 8 layers, with each layer's self-attention and cross-attention having an attention head size ($N_{Head}$) of 4. 
We set the threshold for each score function output to 0.15 for all tasks as the baseline TaskCLIP result. 
Additionally, we provide TaskCLIP's performance with various optimizations in Table~\ref{tab:comparison}. In Section~\ref{sec:ablation}, we will delve into the effects of the transformer aligner module's layer size ($N_{Layer}$), attention head size ($N_{Head}$), and threshold value ($\delta$) on the detection accuracy (AP@0.5).

Among the previous works focusing on task-oriented object detection, TOIST and CoTDet are based on DETR, while the work \cite{sawatzky2019object} employs a pre-trained object detection network as the front-end. However, in the back-end, work~\cite{sawatzky2019object} use Gated Graph Neural Network (GGNN)~\cite{li2015gated} to select objects that could potentially serve as the ideal affordance for specific tasks. Given that both TOIST and GGNN require human involvement to prepare words for specific tasks, they are 	\textbf{task-dependent in an unnatural manner}, making it challenging to apply them to other tasks. Therefore, for a task-agnostic evaluation, we only consider CoTDet and TaskCLIP. To assess CoTDet's performance on the synonym task, we adhere to its ChatGPT prompt and utilize RoBERTa to produce new item and visual feature text embeddings. For TaskCLIP, we reference \cref{tab:LLM_prompt}.(c), requesting ChatGPT to choose visual description words from the visual description word banks maintained during the training phase. These newly selected description words are then directly used to enable TaskCLIP to perform object detection. %It is worth noting that both CoTDet and TaskCLIP were trained exclusively on the original COCO-Tasks dataset images and tasks. Hence, we believe these new tasks can effectively evaluate the models' generalizability.

As indicated by \cref{tab:comparison}, TaskCLIP surpasses both GGNN and TOIST on the original COCO-Tasks dataset. Although TaskCLIP's accuracy is lower than CoTDet's on the original COCO-Tasks dataset, it outperforms CoTDet by over \textbf{9\%} on the new synonyms dataset. This is because CoTDet's training involves direct training of descriptive words and affordance text embedding vectors. Consequently, the accuracy of CoTDet significantly declines when new tasks arise and ChatGPT changes its responses. Conversely, TaskCLIP concentrates on learning the visual features of tasks, which results in better generalizability. 

% TaskCLIP achieves a 4.5\% higher mAP@0.5 than the state-of-the-art (SOTA) model, CoTDet~\cite{tang2023cotdet} (our own implementation version), without employing knowledge distillation. 
% It's worth mentioning that the accuracy reported for CoTDet in Table~\ref{tab:comparison} is based on our own implementation version.
% Compared to TOIST (using noun-pronoun distillation), TaskCLIP achieves a 3.3\% improvement in mAP@0.5 accuracy. 

As shown in Table~\ref{tab:model_compare}, TaskCLIP's training efficiency surpasses that of TOIST and CoTDet due to TaskCLIP's utilization of existing pre-trained vision-semantic information rather than recalibrating them from scratch.

\subsection{Ablation Study} \label{sec:ablation}
\subsubsection{Hyperparameter Tuning} \label{sec:hyper}
\begin{table*}
\centering
\caption{Transformer aligner model architecture search.}
\label{tab:decoder_archsearch}
\resizebox{0.95\textwidth}{!}{
\begin{tabular}{c|cccccccccccc|cccccccccccc}
\toprule
 &
  \multicolumn{12}{c|}{threshold = 0.2} &
  \multicolumn{12}{c}{threshold = 0.4} \\ \midrule
 &
  \multicolumn{3}{c|}{$N_{Layer}$ = 4} &
  \multicolumn{3}{c|}{$N_{Layer}$ = 6} &
  \multicolumn{3}{c|}{$N_{Layer}$ = 8} &
  \multicolumn{3}{c|}{$N_{Layer}$ = 10} &
  \multicolumn{3}{c|}{$N_{Layer}$ = 4} &
  \multicolumn{3}{c|}{$N_{Layer}$ = 6} &
  \multicolumn{3}{c|}{$N_{Layer}$ = 8} &
  \multicolumn{3}{c}{$N_{Layer}$ = 10} \\ \midrule
$N_{Head}$ &
  4 &
  8 &
  \multicolumn{1}{c|}{16} &
  4 &
  8 &
  \multicolumn{1}{c|}{16} &
  4 &
  8 &
  \multicolumn{1}{c|}{16} &
  4 &
  8 &
  16 &
  4 &
  8 &
  \multicolumn{1}{c|}{16} &
  4 &
  8 &
  \multicolumn{1}{c|}{16} &
  4 &
  8 &
  \multicolumn{1}{c|}{16} &
  4 &
  8 &
  16 \\ \midrule
mAP@0.5(\%) &
  33.5 &
  43.5 &
  \multicolumn{1}{c|}{42.6} &
  43.6 &
  43.4 &
  \multicolumn{1}{c|}{42.7} &
  {\textbf{44.4}} &
  43.0 &
  \multicolumn{1}{c|}{ \textbf{44.2}} &
  43.4 &
  42.6 &
  {\textbf{44.5}} &
  30.2 &
  42.4 &
  \multicolumn{1}{c|}{41.7} &
  42.7 &
  42.6 &
  \multicolumn{1}{c|}{41.6} &
  { \textbf{43.6}} &
  42.2 &
  \multicolumn{1}{c|}{{\textbf{43.4}}} &
  42.7 &
  41.8 &
  {\textbf{43.8}} \\ \bottomrule
  % \multicolumn{25}{l}{\small *Here we use blue and red color to mark the top 3 highest mAP@0.5 under threshold 0.2 and threshold 0.4.} \\
\end{tabular}}
\end{table*}

In this section, we focus on three hyperparameters: the number of hidden layers of the transformer aligner module ($N_{Layer}$), the number of attention heads in each aligner layer ($N_{Head}$), and the score function threshold ($\delta$).
In Table~\ref{tab:decoder_archsearch}, we present the effects of $N_{Layer}$ and $N_{Head}$ on mAP@0.5 across 14 tasks, under two different $\delta$ values, 0.2 and 0.4. We choose Faster R-CNN as the object detection network. As noted by previous studies~\cite{sawatzky2019object}, the occurrence of suitable affordances for each task in the COCO-Tasks dataset is much lower than that of not ideal affordances, with an average ratio of $\sim 1:15$. To address this, we set the threshold below 0.5 for our ablation study of the transformer aligner architecture. 
Table~\ref{tab:decoder_archsearch} demonstrates that the top three configurations yielding the best accuracy results are (8,4), (8,16), and (10,16) in the format of ($N_{Layer}$,$N_{Head}$). Among these, we believe the combination ($N_{Layer}$: 8, $N_{Head}$: 4) strikes the best balance between model size and accuracy. In the investigation of threshold effects, we consider the data imbalance of COCO-Tasks. Here, we employ g-means~\cite{jain2009supervised} to determine each task's optimal threshold. The average threshold ($\delta$) over all 14 tasks for aligner configuration (8,4) is 0.15, which is also the $\delta$ value used in Table~\ref{tab:comparison}.

\begin{table*}
\centering
\caption{Ablation study of different model components' effect.}
\label{tab:Ablation study}
\resizebox{0.95\textwidth}{!}{
\begin{tabular}{c|cccccccccccccc|c}
\toprule
\multicolumn{1}{c|}{AP@0.5 (\%)} &
  task1 &
  task2 &
  task3 &
  task4 &
  task5 &
  task6 &
  task7 &
  task8 &
  task9 &
  task10 &
  task11 &
  task12 &
  task13 &
  \multicolumn{1}{c|}{task14} &
  Mean(\%) \\ \midrule
\multicolumn{1}{c|}{(a) without aligner} &
  25.3 &
  35.3 &
  30.1 &
  26.5 &
  24.9 &
  21.4 &
  40.3 &
  4.6 &
  16.9 &
  36.2 &
  22.0 &
  22.9 &
  20.7 &
  \multicolumn{1}{c|}{36.7} &
  26.0 (+0.0) \\
\multicolumn{1}{c|}{(b) F-RCNN} &
  48.4 &
  44.6 &
  52.4 &
  60.6 &
  54.4 &
  31.5 &
  67.7 &
  18.6 &
  39.4 &
  55.7 &
  38.1 &
  47.6 &
  46.3 &
  \multicolumn{1}{c|}{67.4} &
  48.1 (+22.1) \\
\multicolumn{1}{c|}{(c) F-RCNN+diff $\delta$} &
  48.4 &
  44.8 &
  52.4 &
  60.6 &
  54.5 &
  31.5 &
  67.6 &
  18.8 &
  39.4 &
  55.9 &
  38.1 &
  47.6 &
  46.3 &
  \multicolumn{1}{c|}{67.4} &
  48.6 (+22.6) \\
\multicolumn{1}{c|}{(d) F-RCNN+SG} &
  52.0 &
  47.4 &
  54.1 &
  62.5 &
  56.7 &
  34.0 &
  69.9 &
  19.2 &
  41.5 &
  58.1 &
  39.8 &
  49.8 &
  48.7 &
  \multicolumn{1}{c|}{69.8} &
  50.1 (+24.1) \\
\multicolumn{1}{c|}{(e) F-RCNN+both} &
  52.1 &
  47.4 &
  54.1 &
  62.5 &
  56.9 &
  34.0 &
  69.9 &
  19.2 &
  41.4 &
  58.1 &
  39.8 &
  49.8 &
  48.7 &
  \multicolumn{1}{c|}{69.8} &
  50.3 (+24.3) \\
\multicolumn{1}{c|}{(f) YOLOv8} &
  48.7 &
  46.6 &
  45.6 &
  57.1 &
  50.1 &
  30.9 &
  62.5 &
  22.0 &
  38.0 &
  53.9 &
  36.9 &
  40.0 &
  42.6 &
  \multicolumn{1}{c|}{65.4} &
  47.5 (+21.5) \\
\multicolumn{1}{c|}{(g) YOLOv8+diff $\delta$} &
  48.7 &
  45.4 &
  45.6 &
  57.1 &
  50.1 &
  31.1 &
  61.0 &
  23.1 &
  38.2 &
  55.6 &
  36.3 &
  40.5 &
  43.1 &
  \multicolumn{1}{c|}{65.3} &
  47.7 (+21.7) \\
\multicolumn{1}{c|}{(h) YOLOv8+SG} &
  48.6 &
  46.6 &
  45.9 &
  57.1 &
  50.4 &
  31.1 &
  62.5 &
  22.2 &
  38.3 &
  54.3 &
  37.1 &
  40.0 &
  42.6 &
  \multicolumn{1}{c|}{64.4} &
  47.8 (+21.8) \\
\multicolumn{1}{c|}{(i) YOLOv8+both} &
  49.0 &
  46.2 &
  46.0 &
  57.5 &
  50.4 &
  31.3 &
  61.5 &
  23.2 &
  38.5 &
  55.6 &
  36.6 &
  40.5 &
  43.3 &
  \multicolumn{1}{c|}{66.9} &
  49.0 +(23.0) \\ \midrule
\multicolumn{1}{c|}{(j) ground truth bounding box} &
  61.2 &
  73.0 &
  56.3 &
  76.6 &
  56.6 &
  47.4 &
  81.5 &
  28.7 &
  49.9 &
  67.7 &
  47.9 &
  56.0 &
  52.1 &
  \multicolumn{1}{c|}{81.5} &
  62.5 \\ \bottomrule
\multicolumn{16}{l}{\small *Here we use F R-CNN to represent Faster R-CNN.} \\
\end{tabular}}
\vspace{-2em}
\end{table*}

\subsubsection{TaskCLIP Component Analysis} \label{sec:component}
Table~\ref{tab:Ablation study} illustrates the contribution of each component of TaskCLIP to the final accuracy, including the object detection network, transformer aligner module, and select-by-grouping mechanism. 
We compare the performance with the baseline, which directly selects the objects based on the affinity matrix without any re-calibration and alignment. For assessing the effect of the transformer aligner, we present the accuracy of the combination of OpenCLIP and the score function ($f_{score}$). The original OpenAI CLIP and OpenCLIP models were primarily designed to match images with nouns, making it challenging for them to accurately match each bounding box item image with its corresponding visual feature attributes. %As depicted in Table~\ref{tab:Ablation study} (c) and (g), the mAP@0.5 improves by approximately 20\%.
For Table~\ref{tab:Ablation study} (b-e) and Table~\ref{tab:Ablation study} (f-i), we depict the effects of various optimizations on TaskCLIP's final performance. %Specifically, Table~\ref{tab:Ablation study} (b-e) utilizes Faster R-CNN, while Table~\ref{tab:Ablation study} (f-i) employs YOLOv8.
%In Table~\ref{tab:Ablation study} (j), we present the AP@0.5 results obtained when using the ground truth bounding boxes.
%We conduct two pairs of comparisons, namely (b) and (c), (f) and (g), to illustrate the effect of the different threshold strategy (diff $\delta$). In (b) and (f), we utilize the average threshold of 0.15, while in (c) and (g), we apply different thresholds based on g-means for each task. % The accuracy improvement achieved with different $\delta$ values is 0.1\% and 0.2\% when using Faster R-CNN and YOLOv8 as the object detection network, respectively.
%In Table~\ref{tab:Ablation study}.(d) and Table~\ref{tab:Ablation study}.(h), we present the accuracy improvement when using select-by-grouping mechanism ($SG$). Compared to normal setting, $SG$ mechanism provide 2.0\% and 0.3\% accuracy improvement when using Faster R-CNN and YOLOv8 as the detection network. 
%In Table~\ref{tab:Ablation study} (e) and Table~\ref{tab:Ablation study} (i), we present the final AP@0.5 results obtained when deploying both optimization techniques. With the implementation of both optimizations, TaskCLIP achieves a 24.3\% and 23.0\% improvement in mAP@0.5 compared to the baseline (Table~\ref{tab:Ablation study} (a)). 
TaskCLIP effectively recalibrates the item embeddings (vision space) with task-related attribute embeddings (text embedding). We report the accuracy obtained with ground truth bounding boxes in Table~\ref{tab:Ablation study} (j).
%However, when compared to using ground truth bounding boxes (Table~\ref{tab:Ablation study} (j)), TaskCLIP experiences approximately a 12.2\% decrease in mAP@0.5. We attribute this drop in accuracy primarily to limitations inherent in the detection network. As we will elaborate in section~\ref{sec:visualize}, the bounding boxes generated by Faster R-CNN and YOLOv8 exhibit deviations when compared to the ground truth.

\subsection{Visualization and Discussion} \label{sec:visualize}

\begin{figure}[!t] 
\centering
\includegraphics[width=0.85\columnwidth]{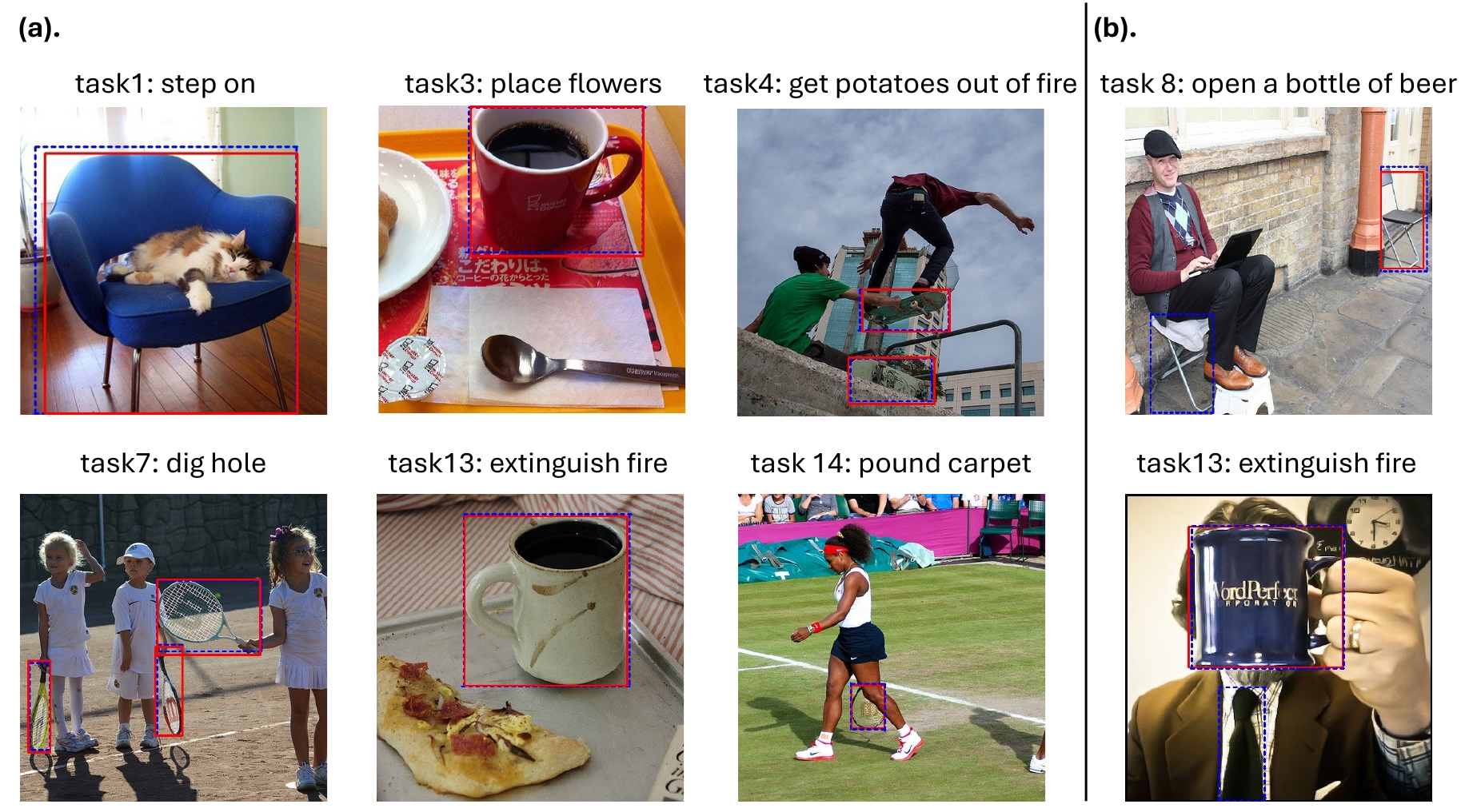}
\caption{Visualization for prediction results of the TaskCLIP (dash blue rectangle) and ground truth (solid red rectangle) (a). Examples with good performance. (b). Examples of unsatisfactory performance. } \label{fig:visualization}
\vspace{-1em}
\end{figure}

Fig.~\ref{fig:visualization} presents the predictions of our model alongside the ground truth in various image samples. 
Specifically, Fig.~\ref{fig:visualization} (a) showcases examples where our model performs well. 
The pridictions exhibit great overlap with the ground truth across various tasks, demonstrating the robust generalizability of our model. However, Fig.~\ref{fig:visualization} (b) highlights two examples where our model's performance is sub-optimal. 
These examples illustrate two common challenges faced by our model. 
Firstly, in the upper panel of Fig.~\ref{fig:visualization} (b), there are two chairs depicted, but only the distant chair can be utilized to open a bottle of beer due to a person occupying the nearby chair. 
While the ground truth accurately captures this subtle distinction, our model erroneously identifies all objects resembling chairs based solely on visual features.
Furthermore, in the lower panel of Fig.~\ref{fig:visualization} (b), our model erroneously selects both a cup and a tie to extinguish fire. 
This misprediction arises from the tie's visual features closely resembling those provided by LLM for solving this task, such as a sturdy handle and a bucket shape. Consequently, our model mistakenly identifies the tie due to its misleading appearance.

\section{Conclusion}
In this study, we introduce TaskCLIP, a novel framework for task-oriented object detection. TaskCLIP leverages pre-trained knowledge and vision-language associations from the frozen CLIP model in an efficient manner, distinguishing itself from prior research efforts. %To address the challenge of misalignment between object image embeddings and their corresponding visual attributes, we propose a transformer-based aligner to recalibrate the vision and text embedding space. Additionally, we propose a select-by-grouping mechanism to mitigate the issue of high false negative mispredictions stemming from imbalanced training data. This mechanism efficiently utilizes the classification output of the object detection network. 
% We validate the efficiency of TaskCLIP through empirical experiments, achieving SOTA performance on the COCO-Tasks dataset. 
Comparative analysis with prior DETR-based approaches demonstrates TaskCLIP's superiority in terms of both task generability, accuracy, and training efficiency.

\section*{Acknowledgements}
This work was supported in part by the DARPA Young Faculty
Award, the National Science Foundation (NSF) under Grants \#2127780,
\#2319198, \#2321840, \#2312517, and \#2235472, the Semiconductor
Research Corporation (SRC), the Office of Naval Research through
the Young Investigator Program Award, and Grants \#N00014-21-1-2225
and N00014-22-1-2067. Additionally, support was provided by the Air
Force Office of Scientific Research under Award \#FA9550-22-1-0253,
along with generous gifts from Xilinx and Cisco.

%\clearpage\mbox{}Page \thepage\ of the manuscript.
% \clearpage\mbox{}Page \thepage\ of the manuscript.
% \clearpage\mbox{}Page \thepage\ of the manuscript.
% \clearpage\mbox{}Page \thepage\ of the manuscript.
% \clearpage\mbox{}Page \thepage\ of the manuscript. This is the last page.
% Now we have reached the maximum length of an ECCV \ECCVyear{} submission (excluding references).
% References should start immediately after the main text, but can continue past p.\ 14 if needed.
% \clearpage  % TODO REVIEW/FINAL: This \clearpage needs to be removed from both review and camera-ready versions.

\bibliographystyle{splncs04}
\bibliography{egbib}
\end{document}